\documentclass[letterpaper, 10 pt, conference]{ieeeconf}

\IEEEoverridecommandlockouts  
\usepackage{graphics} 
\usepackage{graphicx}
\usepackage{amsmath} 
\usepackage{amssymb}  
\usepackage[table,xcdraw]{xcolor}
\usepackage{epsfig}
\usepackage{xcolor}
\usepackage{listings}
\usepackage{verbatim}
\usepackage{multirow}
\usepackage{cite}
\usepackage{algorithm}
\usepackage{algpseudocode}
\usepackage{bm}
\usepackage{booktabs}
\usepackage{siunitx}
\usepackage[table,xcdraw]{xcolor}
\usepackage{subcaption} 

\usepackage{kotex}
\makeatletter
\let\NAT@parse\undefined
\makeatother
\usepackage{hyperref}

\title{\LARGE \bf LiPo: A Lightweight Post-optimization Framework for Smoothing Action Chunks Generated by Learned Policies}

\author{Dongwoo Son$^{1}$, and Suhan Park$^{1}$
\thanks{$^{1}$Dongwoo Son and Suhan Park is with School of Robotics, Colleage of AI Convergence, Kwangwoon University, Republic of Korea.
        {\tt\small ehddnths@kw.ac.kr, park94@kw.ac.kr}}%
}

\begin{document}

\thispagestyle{empty}
\pagestyle{empty}

\maketitle

\begin{abstract}
Recent advances in imitation learning have enabled robots to perform increasingly complex manipulation tasks in unstructured environments. However, most learned policies rely on discrete action chunking, which introduces discontinuities at chunk boundaries. These discontinuities degrade motion quality and are particularly problematic in dynamic tasks such as throwing or lifting heavy objects, where smooth trajectories are critical for momentum transfer and system stability.
In this work, we present a lightweight post-optimization framework for smoothing chunked action sequences. Our method combines three key components: (1) inference-aware chunk scheduling to proactively generate overlapping chunks and avoid pauses from inference delays; (2) linear blending in the overlap region to reduce abrupt transitions; and (3) jerk-minimizing trajectory optimization constrained within a bounded perturbation space. The proposed method was validated on a position-controlled robotic arm performing dynamic manipulation tasks. Experimental results demonstrate that our approach significantly reduces vibration and motion jitter, leading to smoother execution and improved mechanical robustness. Project page: \url{https://sites.google.com/view/action-lipo}
\end{abstract}

\section{INTRODUCTION}

Recent advances in imitation learning (IL) \cite{osa2018algorithmic} and reinforcement learning (RL) \cite{peng2018deepmimic} have enabled robotic systems to autonomously adapt to their environments and perform a wide range of complex tasks. These include simple pick-and-place operations and more intricate tasks, such as folding laundry or manipulating deformable objects, which are notoriously difficult to program through traditional rule-based methods because of the various states of the objects.

In particular, the introduction of action chunking with transformers (ACT) \cite{zhao2023actaloha} has significantly improved the ability of robots to handle long-horizon, deformable, and contact-rich tasks. Impressive results have been demonstrated on tasks such as sliding a Ziploc bag, slotting a battery, opening a cup lid, threading Velcro, preparing tape, and putting on a shoe. These tasks are challenging due to the variability in object states and the difficulty of specifying precise rules or trajectories in advance.

\begin{figure}[t]
    \centering
    \includegraphics[width=\linewidth]{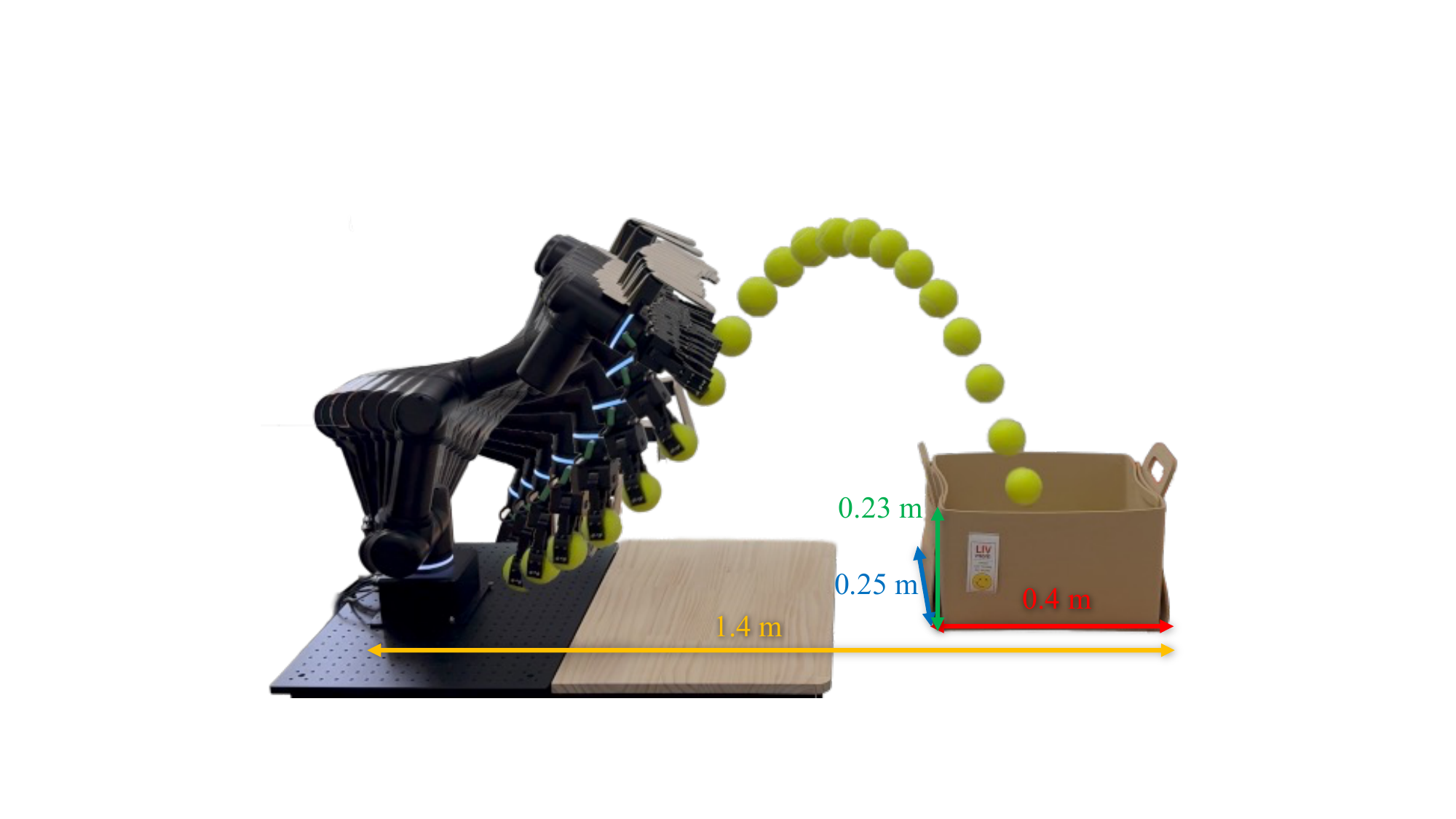}
    \caption{A dynamic throwing motion in which the robotic arm uses momentum to toss a ball into a basket. The trajectory requires smooth and continuous control to ensure accurate release and effective use of inertial forces.}
    \label{fig:throwing_intro}
    \vskip -0.5pc
\end{figure}

Despite these successes, most existing methods operate in relatively static environments and at slow execution speeds. In contrast, real-world robotic systems are increasingly required to perform dynamic motions such as throwing, flipping, or lifting heavy objects, as illustrated in Fig. \ref{fig:throwing_intro}. These tasks demand smooth and temporally precise trajectories to effectively transfer momentum and maintain physical plausibility. However, policies learned from IL or RL often exhibit discontinuities in action sequences. These discontinuities can lead to unstable or jerky execution, which is problematic for dynamic tasks and can degrade performance or even damage hardware.

We propose a simple, lightweight post-optimization framework that improves motion continuity by applying jerk-minimizing smoothing between action chunks. We focus on the following three practical considerations observed in real-world deployments:
\begin{itemize}\setlength{\itemindent}{0.5pc}
\item[1)] \textbf{Inference-induced time delay.} In many real-world demonstration scenarios, robots frequently pause between executing action chunks due to inference latency. This occurs because a new chunk is computed only after the previous one has been fully executed. Such stop-and-go behavior is unacceptable for dynamic tasks. To mitigate this, we proactively generate the next chunk before the current one completes and create an overlapping region to enable seamless transitions.
\item[2)] \textbf{Blending zone.} In the overlapping region, we perform linear interpolation between adjacent chunks to create a first-order smooth transition. This reduces sudden jumps in position or velocity between chunks and provides a better initialization for further optimization.
\item[3)] \textbf{Jerk minimization.} Finally, we refine the blended trajectory using a jerk-minimizing optimizer. This step smooths the concatenated action sequence while respecting a bounded perturbation limit. We pay particular attention to the blending zone, where trajectory variation is most pronounced, to ensure that the final motion remains both smooth and physically feasible.
\end{itemize}

Our contributions are threefold:
We propose a fast and lightweight method for generating physically smooth trajectories by leveraging the structure of chunked action outputs.
To address discontinuities between action chunks, we introduce a blending zone that ensures seamless transitions and smooth motion continuity.
We empirically demonstrate the effectiveness of our method in reducing vibration in position-controlled robots through real-world experiments.

\section{RELATED WORK}
\subsection{Behavior Cloning}

IL enables robots to acquire complex skills directly from expert demonstrations without requiring hand-designed control policies. Behavior cloning (BC)~\cite{osa2018algorithmic} is a type of IL where an agent learns to perform a task by mimicking expert demonstrations. It treats the problem as a supervised learning task, in which the input is the observed state, and the output is the action of the expert.

Given a dataset of expert demonstrations \( \mathcal{D} = \{(s_i, a_i)\}_{i=1}^N \), where \( s_i \) is a state and \( a_i \) is the corresponding expert action, the goal is to learn a policy \( \pi_\theta(a|s) \) that minimizes the difference between the predicted and expert action. This is typically done by minimizing a loss function such as:
\begin{align}
\mathcal{L}(\theta) = \frac{1}{N} \sum_{i=1}^N \| \pi_\theta(s_i) - a_i \|^2.
\end{align}
Recent methods address motion instability and compounding error by leveraging action chunking, where short sequences of actions are predicted and executed instead of single-step control. Given observations \(o_{t-k_o:t}\), the policy predicts a sequence of \( k_a \) actions:
\begin{align}
\pi_\theta(a_{t:t+k_a} \mid o_{t-k_o:t}).
\end{align}
Transformer-based models, such as ACT~\cite{zhao2023actaloha, fu2024mobile, buamanee2024biact, lee2024interact}, generate action chunks conditioned on observations.
Alternatively, Diffusion Policy~\cite{chi2024universal, chi2024diffusionpolicy} employs denoising diffusion processes to generate action chunks, capturing multi-modal distributions inherent in human demonstrations.

Nevertheless, a key limitation remains: transitions between chunks are rarely smoothed, leading to jerkiness or instability, especially problematic in high-speed tasks. In~\cite{ha2024umionlegs}, even successful tossing behavior suffered from mid-motion inference delays that disrupted trajectory continuity.

Temporal ensemble (TE) has been proposed to improve motion smoothness and incorporate feedback continuously.
The policy is queried at every timestep, and overlapping action predictions are averaged with an exponential weighting scheme: 
\begin{align}
a_t = \frac{\sum_{i} w_i A_t[i]}{\sum_{i} w_i}, \quad w_i = \exp(-m \cdot i).
\end{align}
However, this approach requires policy inference at every timestep, resulting in significant computational overhead. More critically, for multi-modal policies, the weighted averaging process can lead to undesirable behavior by blending distinct action modes into ambiguous or physically infeasible trajectories.
This highlights the need for alternative post-processing strategies that can ensure smooth execution while preserving the distinctiveness and coherence of learned action chunks.

\subsection{Trajectory Refinement and Minimum Jerk Trajectory}
Trajectory refinement has been extensively studied, particularly in the motion planning domain. Optimization-based methods such as STOMP~\cite{kalakrishnan2011stomp} and CHOMP~\cite{zucker2013chomp} have been widely adopted to generate smooth and collision-free trajectories by minimizing cost functions related to motion smoothness and safety. Probabilistic approaches, including the use of Gaussian Mixture Models (GMMs) and regression techniques, have also been employed to learn and generate smooth motion patterns from demonstrations~\cite{calinon2007learning}.

Another well-established line of work focuses on generating smooth trajectories by minimizing jerk. Minimum jerk trajectory generation has been widely used to improve smoothness, safety, energy efficiency, and mechanical durability in robotic systems~\cite{kyriakopoulos1994minimum, piazzi2000global, panfeng2006globalminjerk}. In addition to direct minimization, jerk constraints have been integrated into time-optimal trajectory planning frameworks to enhance dynamic feasibility and execution safety~\cite{liu2016smooth, lee2024jerk}.

While these methods effectively generate smooth motions, they incur significant computational cost, limiting their use in real-time or dynamic execution scenarios. To address this gap, we propose a lightweight post-optimization framework that bridges discrete learned action chunks and the continuous demands of physical systems. Our approach enforces trajectory smoothness across chunk boundaries while accounting for inference delay and bounded deviation constraints, to enable reliable execution in dynamic tasks. The following section details this method.

\section{LIGHTWEIGHT POST-OPTIMIZATION}
We propose a two-stage post-optimization framework to achieve smooth transitions between discrete action chunks in dynamic settings. The key idea is to locally refine the learned trajectory using a lightweight optimization process that respects inference delay and bounded deviation constraints. First, we perform a linear blending between overlapping chunks to provide an initial smooth transition. Then, we apply a jerk-minimizing refinement to further enhance continuity and mechanical stability. The following subsections describe each step in detail.

\begin{figure}
    \centering
    \includegraphics[width=1\linewidth]{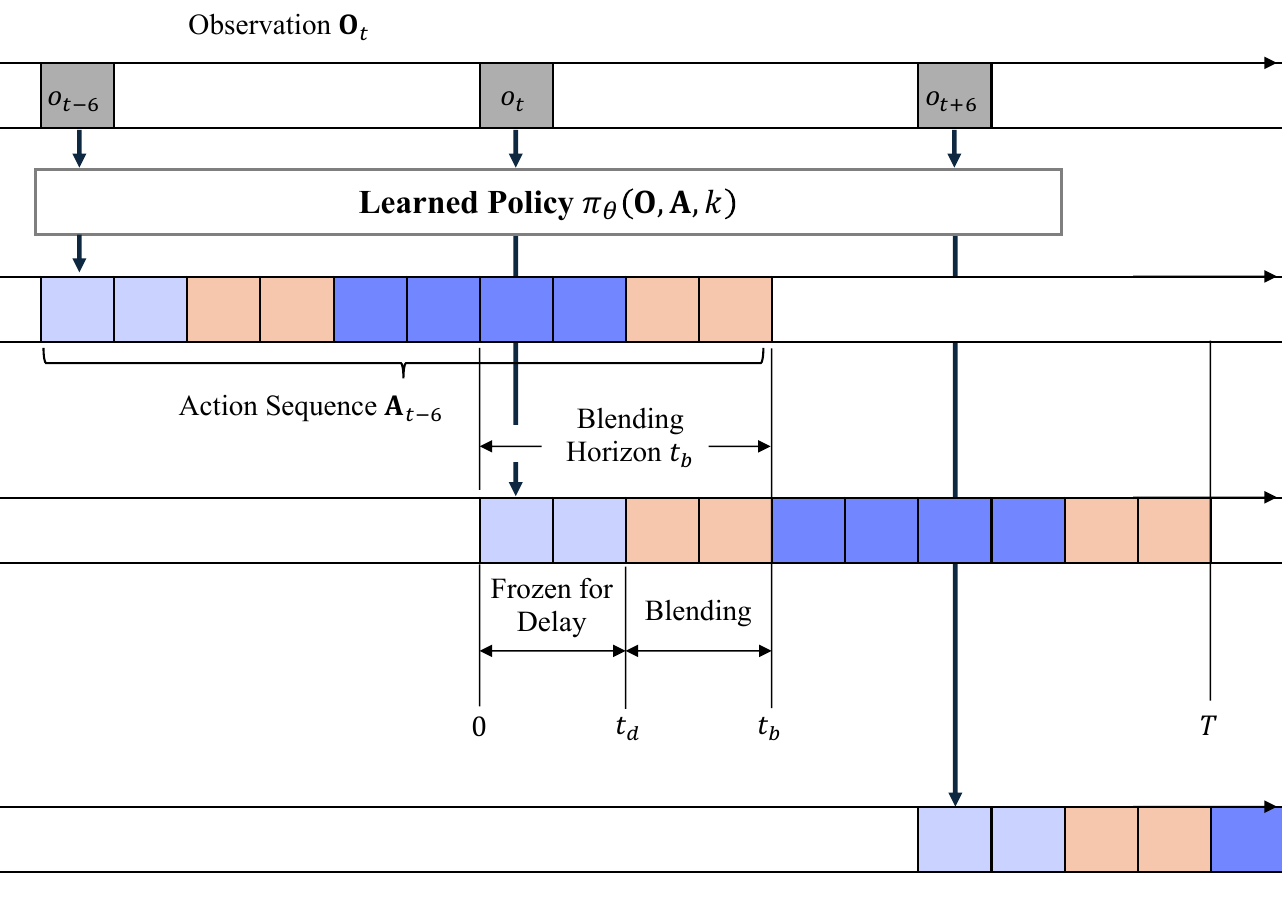}
    \caption{Concept of reference trajectory construction with time delay and linear blending.
    The shaded orange zone represents the linear interpolation area, while the shaded light blue region indicates the inference delay window where no blending occurs.
   }
    \label{fig:chunk_overlap}
\end{figure}
\subsection{Reference Trajectory Construction with Inference Time Delay}

A linearly blended reference trajectory \( q_{\text{ref}}(t) \) is constructed to interpolate between the previously optimized and newly predicted action chunks. This blending accounts for inference-induced time delay by freezing the transition during the delay window and then smoothly interpolating over the defined blending region.

The linearly blended reference trajectory is defined as:
\begin{align}
q_{\text{ref}}(t) =
\begin{cases}
q_{\text{past}}(t), & t \in [0, t_d] \\
    (1 - \alpha(t)) q_{\text{past}}(t) +  \\ \quad \quad \alpha(t) q_{\text{new}}(t), & t \in (t_d, t_b] \\
q_{\text{new}}(t), & t \in (t_b, T]
\end{cases},
\end{align}
where the blending weight function \( \alpha(t) \) is given by:
\begin{align}
\alpha(t) = \frac{t - t_d}{t_b - t_d},
\end{align}
and \( q_{\text{past}}(t) \) is the trajectory from the previous action chunk,  
\( q_{\text{new}}(t) \) is the newly inferred action chunk,  
\( t_d \) denotes the time delay caused by inference,  
\( t_b \) is the end of the blending zone,  
and \( T \) is the total length of the action segment.
Fig. \ref{fig:chunk_overlap} illustrates the detailed concept (See Fig. \ref{fig:lipo_result_plot} for the detailed real-world optimization results). Note that we use $q$ to denote actions instead of $a$ in this section for consistency with trajectory optimization literature.

\subsection{Minimum Jerk Chunk Post-optimization}

To further smooth the sequence of action chunks and ensure safe, physically consistent execution, we introduce a post-optimization step that minimizes the overall jerk of the trajectory. Given a reference trajectory \( q_\mathrm{ref}(t) \) composed of discrete action chunks, we optimize a small perturbation \( \epsilon(t) \) such that the resulting trajectory \( q_\mathrm{ref}(t) + \epsilon(t) \) minimizes jerk:
\begin{align}
\min_{\mathbf{\epsilon}} \quad & \int_{0}^{T} \left\| \frac{d^3}{dt^3}(q_\mathrm{ref}(t) + \epsilon(t)) \right\|^2 \, dt ,₩&\label{eq:motion_planning} \\
\text{s.t.} \quad &  \epsilon(t)=\mathbf{0}, \quad & \forall t \in [0, t_d] \nonumber \\
& \left \| \epsilon(t) \right \|_\infty \leq \bar{\epsilon}_{b}, \quad & \forall t \in (t_d, t_b] \nonumber \\
& \left \| \epsilon(t) \right \|_\infty \leq \bar{\epsilon}_{p}, \quad &\forall t \in (t_b, T] \nonumber 
\end{align}
where $\bar{\epsilon}_{b}$ and $\bar{\epsilon}_{p}$ denote blending and path bounds, respectively. A higher blending bound $\bar{\epsilon}_{b}$ is employed compared to the path bound $\bar{\epsilon}_{p}$, because the chunk boundaries typically involve greater uncertainty during policy transitions.

This formulation encourages smooth transitions across chunk boundaries by penalizing high third-order derivatives, while constraining the deviation \( \epsilon(t) \) to remain within a specified bound. The constraint ensures that the perturbed trajectory remains close to the original, thus preserving the intent and feasibility of the task. 

\subsection{Interpolation for High-frequency Control}

To ensure smooth execution in high-frequency position control systems, we employ quintic spline interpolation between low-rate action samples. Quintic polynomials offer continuity up to the second derivative, making them well-suited for applications requiring smooth velocity and acceleration profiles.

This interpolation provides smoothness across segments and avoids undesirable oscillations in the control signal, which is particularly important when operating at high control frequencies. It also improves the execution fidelity of learned trajectories and reduces mechanical wear caused by discontinuous commands.

\begin{figure}[t]
    \centering
    \includegraphics[width=0.8\linewidth]{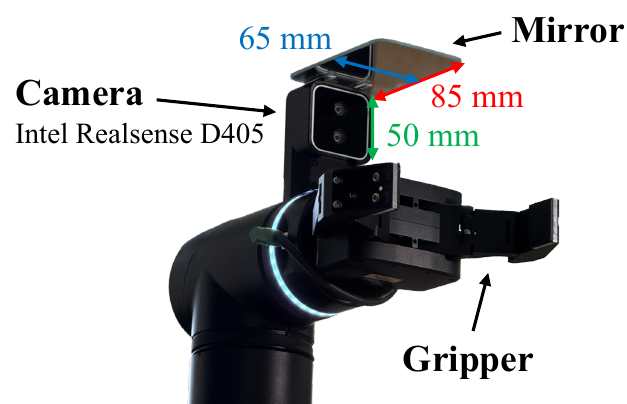}
    \caption{End-effector setup used in the experiments. The red and blue arrows indicate the dimensions of the mirror mounted on the gripper, while the green arrow represents the height of the installation point. The colored labels indicate the lengths corresponding to each arrow.}
    \label{fig:ee_setup}
\end{figure}
\subsection{Task-Space Error Bound from Joint-Space Perturbation}

The proposed post-optimization method perturbs the joint trajectory within a bounded range to smooth transitions between action chunks. To ensure safety, it is important to estimate the corresponding deviation in task space induced by such perturbations.

Let the joint-space perturbation be bounded as:
\begin{align}
\|\epsilon(t)\|_\infty \leq \bar{\epsilon}.
\end{align}
Using a first-order Taylor approximation of the forward kinematics function \( f(q) \), the resulting deviation in task space can be approximated as:
\begin{align}
    \|f(q(t) + \epsilon(t)) - f(q(t))\| 
   & \approx  \|J(q(t)) \epsilon(t)\| \nonumber \\
   & \leq \|J(q(t))\| \cdot \|\epsilon(t)\|,
\end{align}
where \( J(q(t)) \) denotes the Jacobian matrix of the robot at time \( t \).
Therefore, the worst-case task-space deviation at time \( t \) is upper bounded by:
\begin{align}
    \delta x_{\mathrm{max}}(t) = \|J(q(t))\| \cdot \bar{\epsilon}.
\end{align}

This formulation provides a conservative upper bound on task-space deviation caused by joint-space smoothing, offering a principled measure of safety during post-optimization. In practice, the maximum norm of the Jacobian \( \|J(q)\| \) can be estimated using critical configurations such as fully extended postures, where the robot is most sensitive to perturbations.

\begin{figure}[!t]
    \centering
    \begin{subfigure}[t]{0.49\linewidth}
        \includegraphics[width=\linewidth]{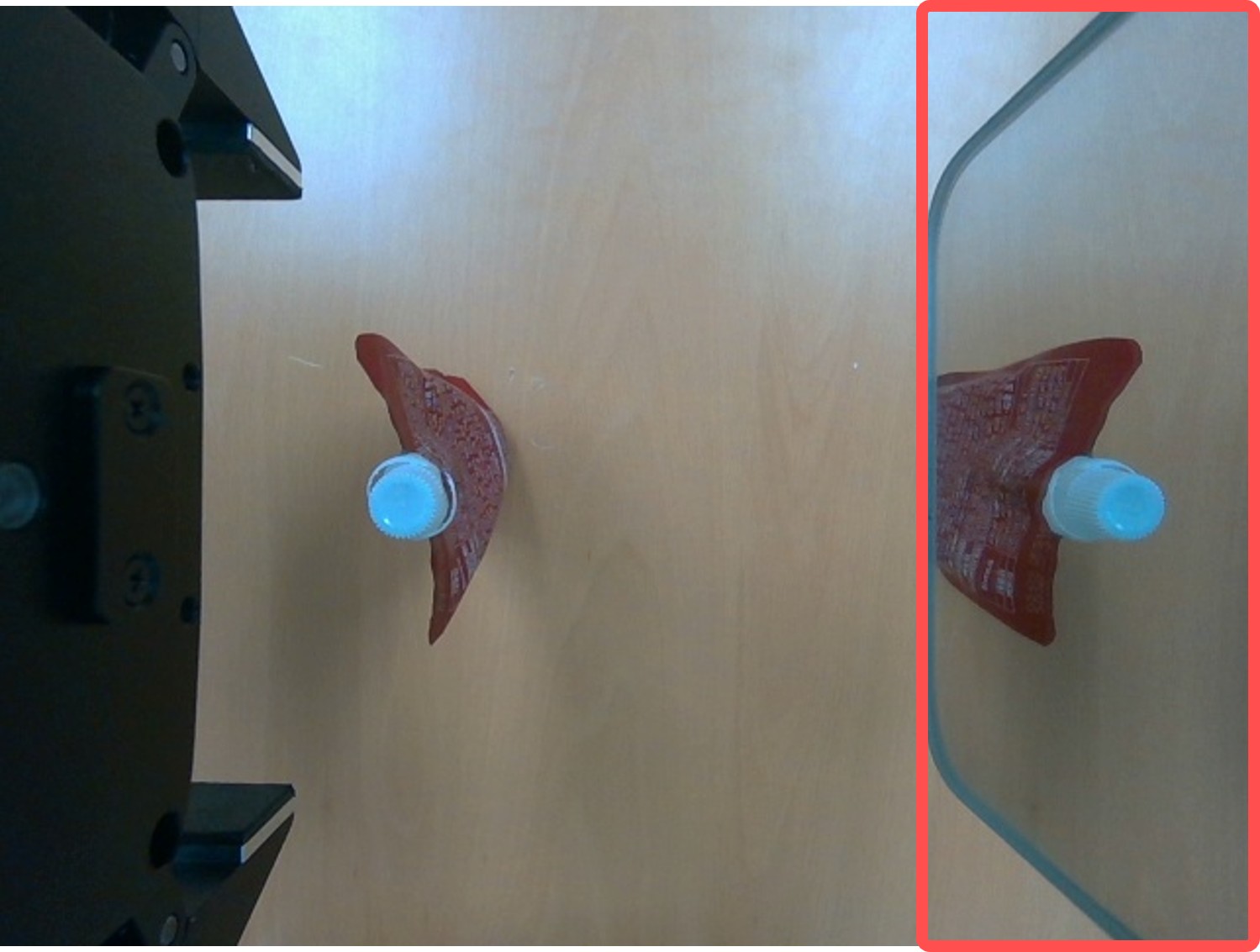}
        \caption{Original image}
        \label{fig:original_image}
    \end{subfigure}
    \hfill
    \begin{subfigure}[t]{0.49\linewidth}
        \includegraphics[width=\linewidth]{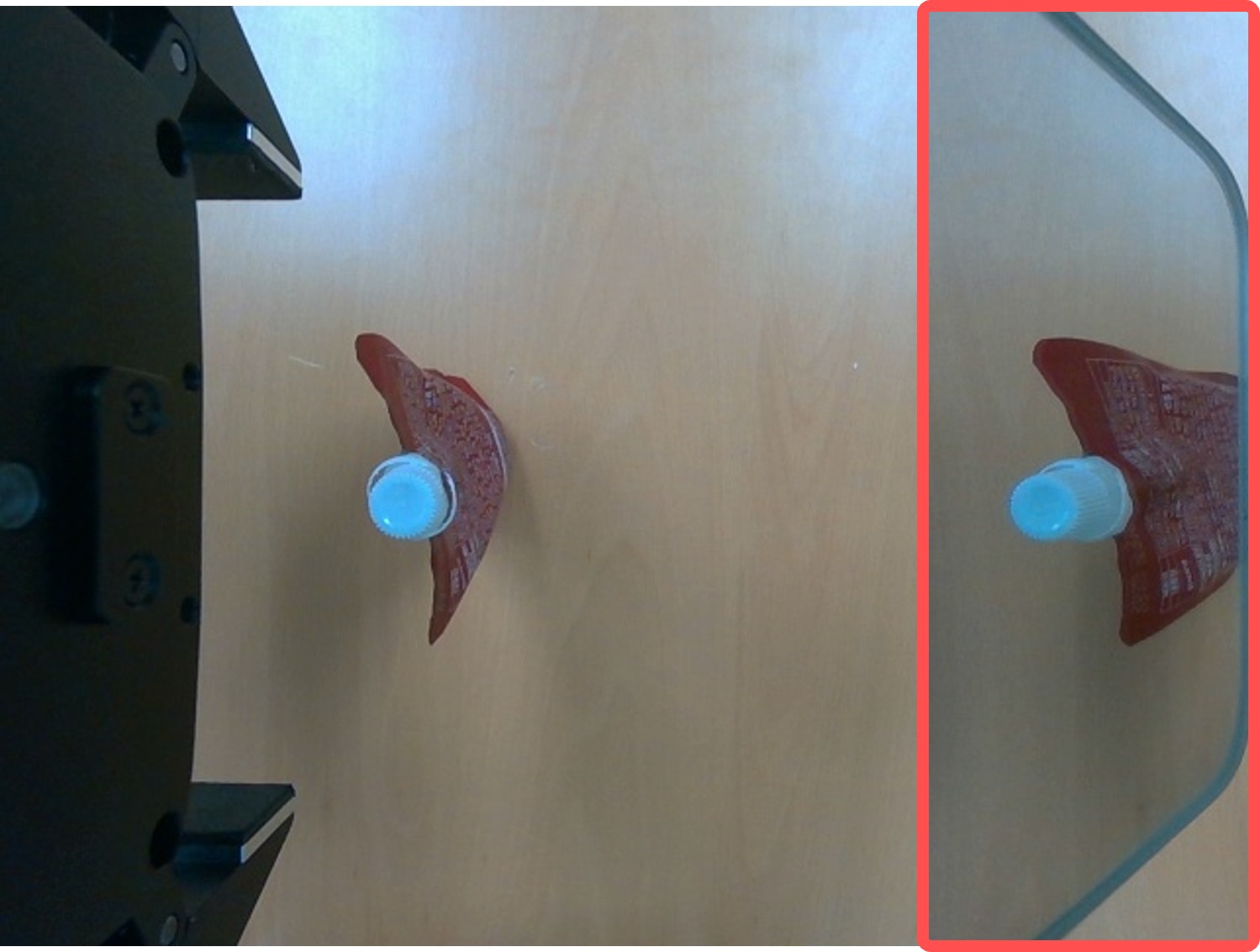}
        \caption{Partially flipped image}
        \label{fig:flipped_image}
    \end{subfigure}
    \vskip -0.5pc
    \caption{Camera input preprocessing: (a) original image and (b) horizontally flipped image using a mirror setup.}
    \label{fig:image_flip}
\end{figure}
\section{EXPERIMENTS}

\subsection{Experimental Setup}
All experiments were conducted using the ROBOTIS OpenManipulator-Y~\cite{robotis_open_manipulator-y} in position control mode. Although the manipulator supports velocity and current control modes, we opted for position control to ensure compatibility with other position-controlled robotic platforms. The control loop was operated at 400~Hz to support high-precision trajectory tracking.

Visual input was captured by a single RGB camera running at 30~Hz with a resolution of 640×480 pixels. As shown in Fig. \ref{fig:ee_setup}, a mirror was placed adjacent to the workspace to provide a secondary viewpoint for implicit stereo perception. The captured image was horizontally flipped, as illustrated in Fig. \ref{fig:image_flip}, to preserve motion linearity between views, which is a similar approach to the UMI gripper system~\cite{chi2024universal}. Policy inference was performed at the same frequency as the camera input, i.e., 30~Hz.

The post-optimization was formulated as a quadratic program and solved using the Clarabel solver~\cite{Clarabel_2024}. $T$, $t_d$, and $t_b$ were set to 1.66 s, 0.33 s, and 0.16 s, respectively, corresponding to 50 total steps for each chunk, a 5-step time delay, and a 10-step blending horizon. 
The perturbation bounds $\epsilon_b$ and $\epsilon_p$ were set to 0.02 rad and 0.003 rad for this experiment.

We used the Lerobot \cite{cadene2024lerobot} platform to evaluate the proposed method in a real-world environment. ACT \cite{zhao2023actaloha} was used to validate the proposed method. All computations were performed on a system equipped with an Intel i7-14700HX CPU, 32~GB of RAM, and an NVIDIA RTX 4060 Max-Q Mobile GPU with 8~GB of VRAM.

\begin{figure}
    \centering
    \includegraphics[width=0.8\linewidth]{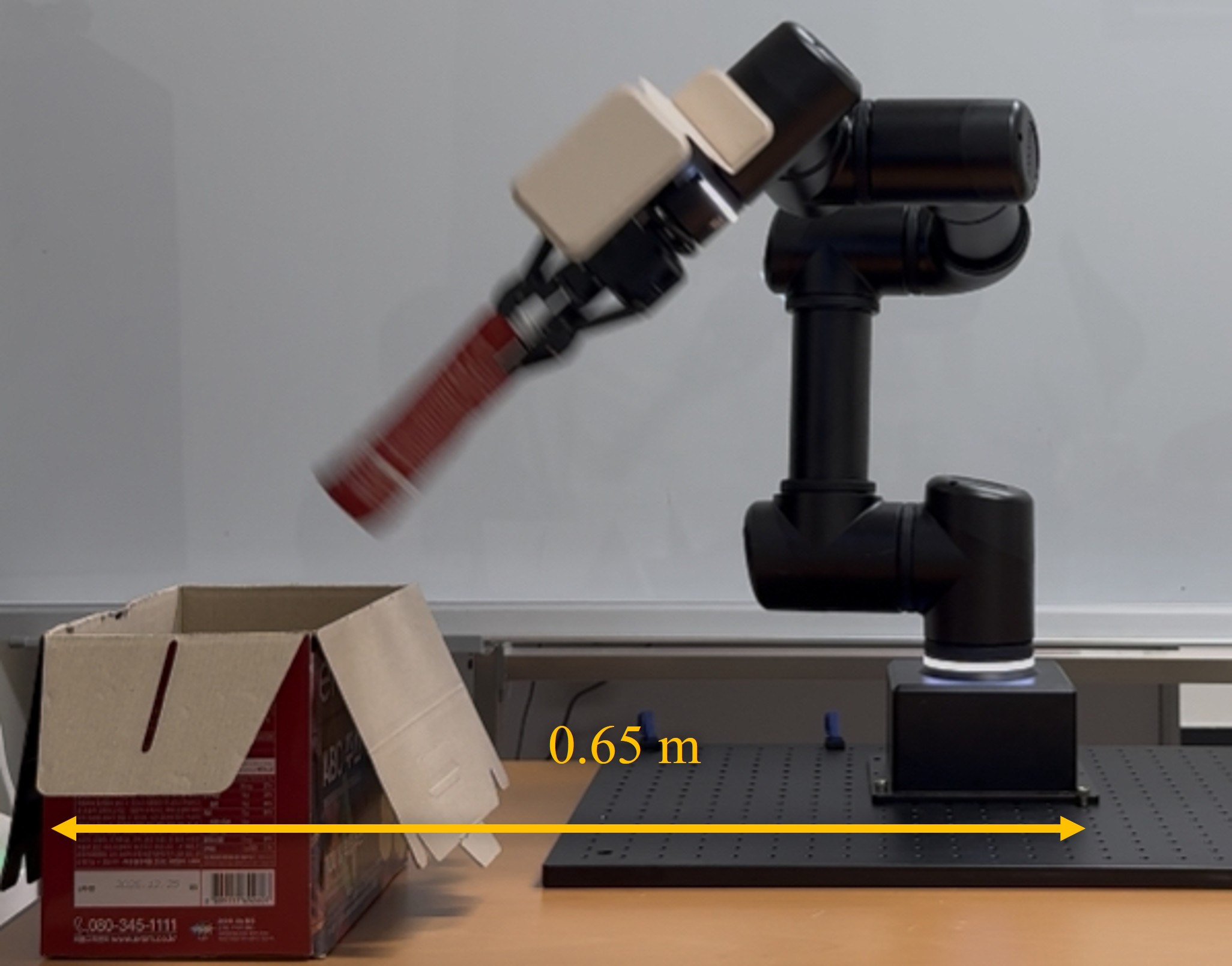}
    \caption{Pouch throw task. The robot grasps a juice pouch and throws it into a box.}
    \label{fig:throwing_pouch}
\end{figure}

\begin{figure*}[t]
    \centering
    \includegraphics[width=0.83\linewidth]{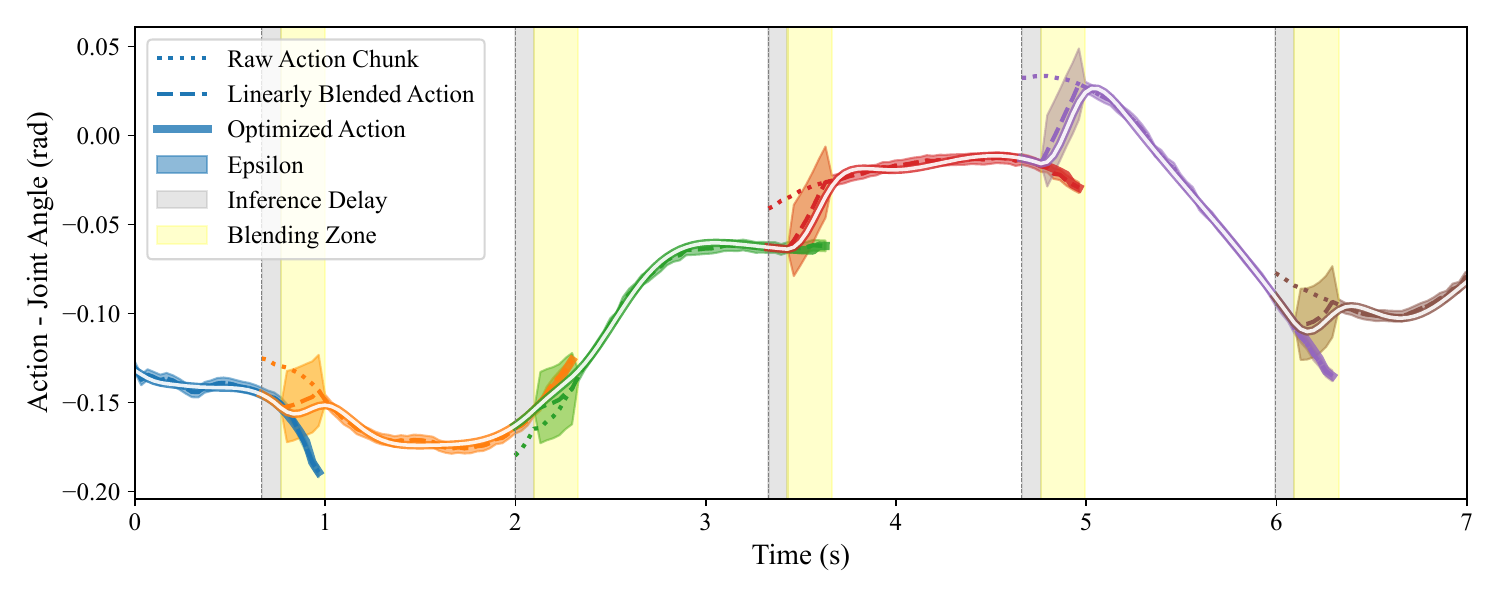}
    \caption{
    Visualization of action chunk smoothing with LiPo.
    Colored segments represent different action chunks predicted by the policy.
    The dotted line shows the raw action trajectory before the post-optimization.
    The dashed line indicates the linearly blended trajectory within overlapping regions.
    Shaded areas highlight the $\epsilon$-constrained perturbation limits.
    The solid line represents the post-optimized chunk trajectory, and the solid white line denotes the final executed trajectory. The yellow shaded areas represent the blending zones where local smoothing is applied.
    }
    \label{fig:lipo_result_plot}
\end{figure*}

\begin{figure*}[t]
    \centering
    \begin{subfigure}[t]{.495\linewidth}
        \centering
        \label{fig:res_w_opt_q}
        \includegraphics[width=0.82\linewidth]{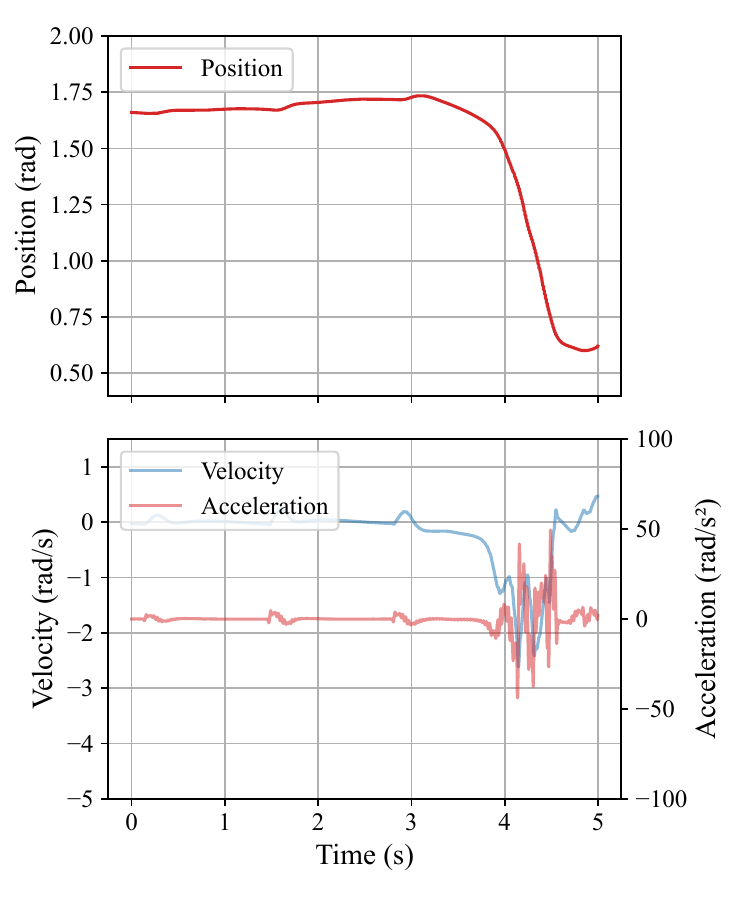}
        \caption{Post-optimized action With LiPo}
    \end{subfigure}
    \begin{subfigure}[t]{0.495\linewidth}
        \centering
        \label{fig:res_wo_opt_q}
        \includegraphics[width=0.82\linewidth]{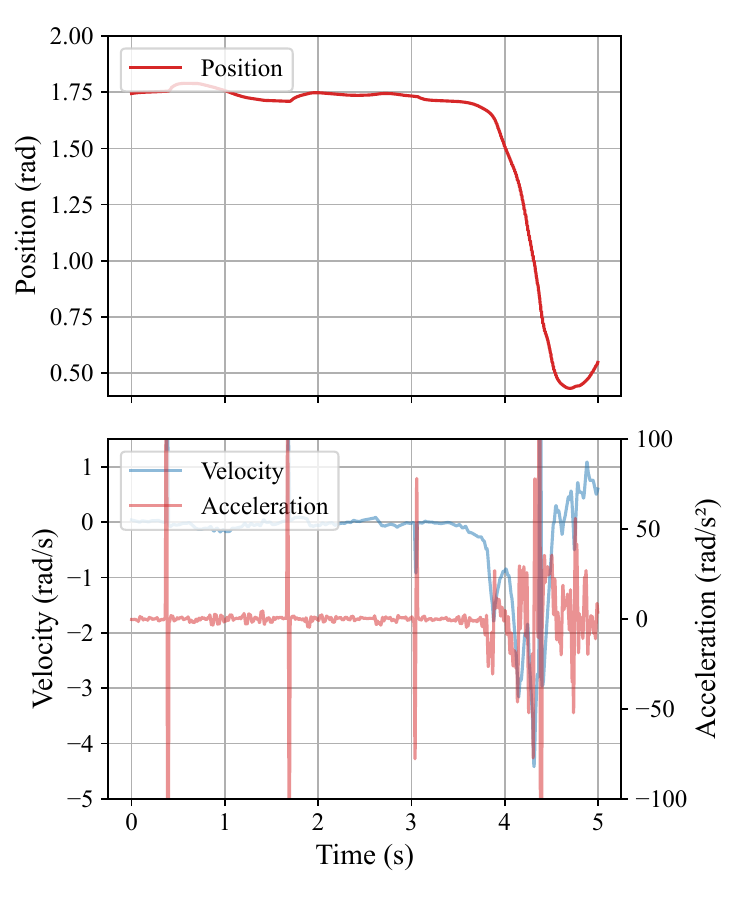}
        \caption{Raw action Without LiPo}
    \end{subfigure}
    \caption{Comparison of joint position, velocity, and acceleration trajectories with and without the proposed LiPo method. LiPo significantly reduces abrupt changes, resulting in smoother and more stable motion.}
    \label{fig:res_comparison}
\end{figure*}

\subsection{Experimental Results}
Two experiments were conducted to evaluate the proposed method: the \textbf{pouch throw} task and the \textbf{ball toss} task, both of which require dynamic robot motions.

In the \textbf{pouch throw }task, illustrated in Fig. \ref {fig:throwing_pouch}, the robot grasps a juice pouch randomly placed within a 10 cm $\times$ 10 cm region on a tabletop and throws it into a box. As shown in the Fig. \ref{fig:lipo_result_plot}, the proposed method smoothly connects the transitions between distinct action chunks, even when the underlying predicted trajectories exhibit noticeable discontinuities. In particular, a smooth trajectory was achieved by reusing the previously optimized trajectory during the time delay period. In the blending region, where model uncertainty tends to be higher, we allow a larger perturbation bound, while enforcing a tighter bound during the remainder of the trajectory to preserve fidelity.
Fig. \ref{fig:res_comparison} shows the joint position, velocity, and acceleration with and without the proposed LiPo method. The proposed method effectively reduces jerky motion, resulting in smoother trajectories.

Fig.~\ref{fig:throwing_intro} illustrates the \textbf{ball toss} task.
In this scenario, the robot detects and grasps a ball on the ground and executes a throwing motion to deliver it into a basket. The task is considered successful if the ball lands inside the basket; otherwise, it is counted as a failure. Unlike the setup in \cite{ha2024umionlegs}, where the ball is pre-grasped at the start, our setting requires the robot to dynamically react and grasp the ball upon placement, introducing real-time perception and response challenges.
This task is particularly difficult for learned policies, as inference latency during execution, when a new action chunk is sampled mid-trajectory, can introduce discontinuities that compromise task success.
Therefore, Smooth and continuous transitions between action chunks are critical for reliable execution in dynamic scenarios like this. We conducted 10 trials per condition, and the results are summarized in Table~\ref{tab:ablation_lipo}.

The experiments show that using LiPo and quintic spline improves task success rates compared to conditions without post-optimization. While linear interpolation after optimization yields visually smoother motions, it often fails to produce dynamically effective throws, highlighting the importance of quintic spline after the optimization for dynamic tasks. We were unable to execute the task using the temporal ensemble (TE) \cite{zhao2023actaloha}, as the robot became unstable during high-speed throwing motions.

\begin{table}[t]
\centering
\caption{Success rate of the ball toss task with and without the proposed method.}
\label{tab:ablation_lipo}
\begin{tabular}{lcc}
\toprule
\textbf{Method} & \textbf{Spline} & \textbf{Success Rate} \\
\midrule
LiPo (Ours)         & Quintic & \textbf{90\%} \\
LiPo (Ours)         & Linear  & 60\% \\
Raw action             & Quintic & 80\% \\
Raw action      & Linear  & 70\% \\
TE & - & - \\
\bottomrule
\end{tabular}
\vskip -1.5pc
\end{table}

\section{DISCUSSION}

\subsection{Optimization Design Choices}
The proposed method adopts an $\ell_\infty$-norm constraint on the perturbation $\epsilon$ to define the feasible optimization region. This decision was made to maintain the lightweight nature of the post-optimization step. By using bound constraints on each variable, the optimization problem becomes a simple box-constrained quadratic program (QP), which significantly improves solver speed and reliability compared to $\ell_1$- or $\ell_2$-norm formulations that require more complex projections or regularization terms.

Notably, we do not impose robot dynamics constraints in the optimization. Although incorporating full robot dynamics (e.g., torque and velocity limits) could theoretically yield more trackable trajectories, this comes at a significant computational cost. Moreover, if the policy itself produces actions that violate the robot's dynamic feasibility, it indicates a failure in policy generation rather than execution. Attempting to retroactively enforce feasibility through trajectory reshaping may not always lead to desirable or correct behavior.

Additionally, incorporating dynamics constraints often necessitates adjusting the trajectory duration to remain within dynamic feasibility bounds. This introduces a temporal mismatch between the optimized trajectory and the timing assumed by the learned policy, potentially undermining the task intent encoded by the demonstrator. To preserve timing consistency and keep the optimization problem tractable, we intentionally exclude dynamics constraints from our formulation.

\section{CONCLUSION}

This paper proposed a lightweight yet effective post-optimization framework to improve the smoothness and physical consistency of chunk-based robot motion trajectories. The proposed method incorporates linear blending across chunk boundaries and minimum-jerk optimization under bounded joint-space perturbations, enabling smooth transitions without altering the structure of the underlying learned policy.

Experiments on the ROBOTIS Manipulator-Y demonstrated that the proposed approach significantly reduces motion discontinuities and mitigates mechanical vibrations, particularly in high-frequency position control settings. Additionally, the use of quintic spline interpolation further enhanced execution fidelity by generating continuous, high-resolution motion profiles suitable for real-time control.

The proposed framework is conceptually simple yet practically robust. It requires minimal computational resources, is broadly applicable across different robot platforms and learning methods, and can be seamlessly integrated into existing imitation or reinforcement learning pipelines. Future work includes extending this framework to handle multi-modal trajectory refinement, contact-aware blending, and real-time closed-loop adaptation under dynamic constraints.

Future work includes extending the framework to handle contact-aware blending. Incorporating torque limits and joint velocity constraints while preserving the task performance is a promising direction for improving physical feasibility.

\section*{ACKNOWLEDGMENTS}
We gratefully acknowledge ROBOTIS for generously providing the Physical AI research package, which included the robotic arm and remote teleoperation platform.

\appendices

\bibliographystyle{IEEEtran}
\bibliography{main}

\end{document}